\documentclass[10pt,twocolumn,letterpaper]{article}

\usepackage{cvpr}
\usepackage{times}
\usepackage{epsfig}
\usepackage{graphicx}
\usepackage{amsmath}
\usepackage{amssymb}
\usepackage[]{algorithm2e}

\usepackage[pagebackref=true,breaklinks=true,letterpaper=true,colorlinks,bookmarks=false]{hyperref}

\cvprfinalcopy 

\def\cvprPaperID{4} 

\ifcvprfinal\fi
\begin{document}

\title{An approach to image denoising using manifold approximation without clean images}

\author{
Rohit Jena \\
Department of Computer Science and Engineering \\
Indian Institute of Technology, Bombay \\
{\tt\small rohitrango@cse.iitb.ac.in}
}

\maketitle

\begin{abstract}
Image restoration has been an extensively researched topic in numerous fields. With the advent of deep learning, a lot of the current algorithms were replaced by algorithms that are more flexible and robust. Deep networks have demonstrated impressive performance in a variety of tasks like blind denoising, image enhancement, deblurring, super-resolution, inpainting, among others. Most of these learning-based algorithms use a large amount of clean data during the training process. However, in certain applications in medical image processing, one may not have access to a large amount of clean data. In this paper, we propose a method for denoising that attempts to learn the denoising process by pushing the noisy data close to the clean data manifold, using only noisy images during training. Furthermore, we use perceptual loss terms and an iterative refinement step to further refine the clean images without losing important features.
\end{abstract}

\section{Introduction}
Image denoising is a problem that has been researched for over the last few decades and is still a very actively research topic due to its vast applications in medical image processing. Noise can creep into the image during the acquisition process, or during processing of the image post-acquisition. Different noise models capture the noisy image generation in different stages of the image acquisition and processing. Over these years, there have been a myriad of denoising algorithms which take into account these noise models, constraints on images, blind v/s non-blind denoising, i.e. knowing the noise model parameters or type. In the recent years, deep learning has proven to be very effective at many tasks, even surpassing humans in some of them \cite{russakovsky2015imagenet}. A variety of techniques have also been applied in image denoising, with various constraints and assumptions. Some famous image denoising methods are discussed in the next section.

\section{Related Work}
Image restoration is a long standing problem with a lot of applications and use-cases. 
Some of the popular algorithms for denoising are \cite{Dabov_undated-dy}, which deals with Gaussian data, \cite{Aharon2006-dd} for sparse reconstruction of signals and \cite{anscombe1948transformation} which is used for Poisson and binomial data. Previous work based on manifold denoising is also done by \cite{NIPS2006_2997} where a graph-based diffusion process is used on the point sample. 
With the advent of deep learning, algorithms have been developed that deal with supervised, semi-supervised and unsupervised image restoration. \cite{Yang2018BM3DNetAC} attempts to unroll the computational pipeline of BM3D algorithm into a convolutional neural network structure. \cite{Vincent:2010:SDA:1756006.1953039} propose a strategy of using autoencoders to denoise corrupted versions of their inputs. \cite{Creswell_2018} takes another step by using adversarial training to shape the latent shape in addition to learning to produce denoised outputs. Some work on denoising using manifolds is also done in \cite{tripathi2018correction}, where they propose to denoise an image by finding the closest point in the manifold of the GAN which is learned on clean images. \cite{NIPS2012_4686} uses deep networks pretrained with denoising autoencoder, for image inpainting and denoising. \cite{xu2018trilateral} use a more complex trilateral weighted sparse coding for real world image denoising.
In medical settings, there has been a lot of work as well. \cite{wang2006total} use a dictionary learning based algorithm for denoising, \cite{wang2016accelerating} proposes a deep learning approach to reconstruct MR data from zero-filled MR images. \cite{chen2017low} use a residual convolutional network for low-dose CT imaging. \cite{gondara2016medical} also tackle the medical image denoising problem using convolutional autoencoders.  

However, all these algorithms use a lot of training data to work. However, getting large amounts of training data, which may not be feasible in medical settings due to large costs of acquiring high-quality data. Certain modalities like X-ray may also be harmful for the patients due to high radiation dose. Hence, this calls for algorithms in unsupervised learning. \cite{ulyanov2018deep} showed that neural networks can be used as very strong priors, and image restoration tasks like blind denoising, inpainting, superresolution, and other tasks can be performed for single image without training any network, and by just using a single noisy instance. Recently, \cite{noise2noise} proposed to learn the statistical properties of noise using a different training strategy - to use noisy pairs to train a network. The input is a noisy image, and the output is another noisy image. In their experiments, the noise parameter of both noisy images are the same. Although they show impressive results, acquiring noisy images with the same noise parameter may not complete feasible in certain applications. In contrast, my method uses a single noisy image and has a ``directional'' learning scheme, i.e. the network has to learn mapping from \textit{more noisy} to \textit{less noisy} examples and not the other way, since the training pairs in this scheme are not symmetrical unlike \cite{noise2noise}. 

\begin{figure}
    \centering
    \includegraphics[width=0.9\linewidth]{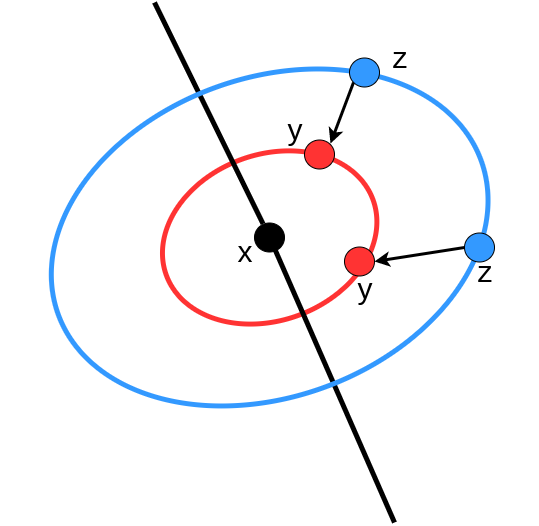}
    \caption{A visualization of the manifold of clean images. The black line represents a low-dimensional manifold of clean images. The clean images $x$ are going to lie on this manifold. The ``contours'' perpendicular to this line is shown as red and blue circles. Instances of $y$ are present in the training set. My method generates instances of $z$ from $y$ (which are more corrupted versions and hence will be farther from the line) and the arrow indicates the mapping the network has to learn, in this case, moving from $z$ to $y$. }
    \label{fig:manifold}
\end{figure}

\section{Method}

\begin{figure*}
\begin{center}
\includegraphics[width=0.15\linewidth]{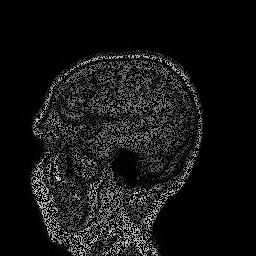}
\includegraphics[width=0.15\linewidth]{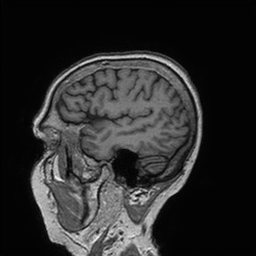}
\includegraphics[width=0.15\linewidth]{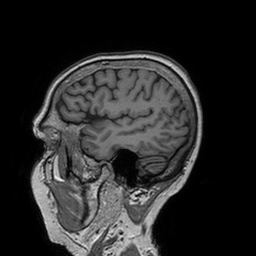}
\hspace{5pt}
\includegraphics[width=0.15\linewidth]{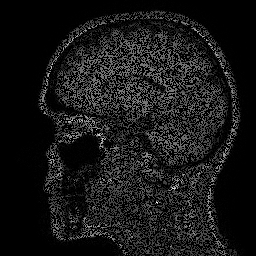}
\includegraphics[width=0.15\linewidth]{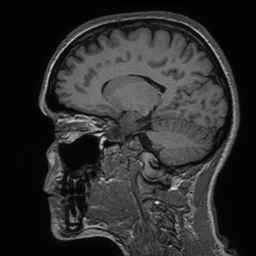}
\includegraphics[width=0.15\linewidth]{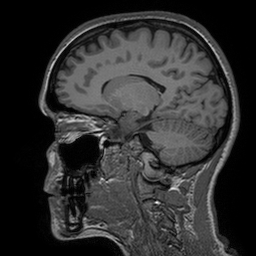} 

\includegraphics[width=0.15\linewidth]{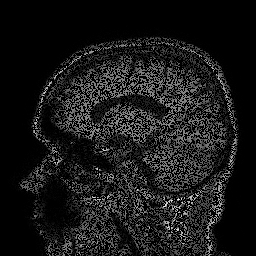}
\includegraphics[width=0.15\linewidth]{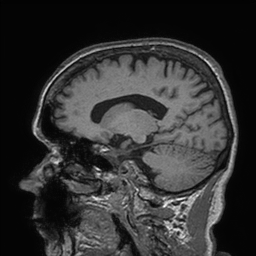}
\includegraphics[width=0.15\linewidth]{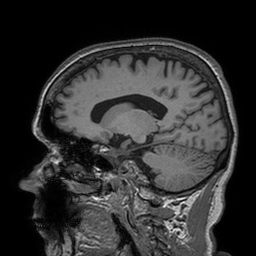}
\hspace{5pt}
\includegraphics[width=0.15\linewidth]{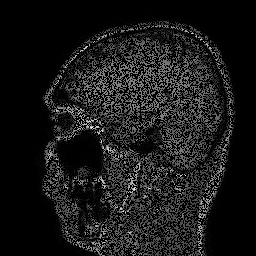}
\includegraphics[width=0.15\linewidth]{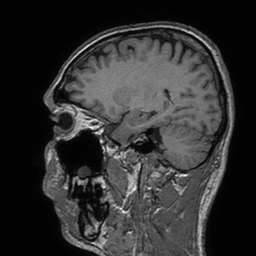}
\includegraphics[width=0.15\linewidth]{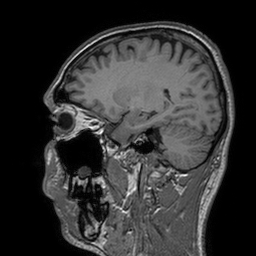}
\end{center}
  \caption{Results on the IXI Guys dataset. The images from left to right are (i) Noisy image (ii) Reconstructed image, and (iii) original image. Note that the network not only fills up the gaps properly, but also retains the edge information and finer details.}
\label{fig:ixiguys}
\end{figure*}

Before we move into the manifold mapping technique, we define some terminology first. Let $M$ be the size of the training set, which consists of only noisy instances\{$y^i$\}, $i = \{1, 2 ... M\}$. Each of the instance $y^i$ corresponds to a clean image $x^i$, from which it is generated via some random process during acquisition, post-processing, etc. More formally,
\begin{equation}
    y^i = D(x^i; \theta^i),  \forall{i} = \{1, 2, ... M\} 
\end{equation}
where $D$ is the degradation model parameterized by noise parameters $\theta^i$. The training set doesn't have the clean images $x^i$ and only contains \textit{one} noisy instance of the corrupt image $y^i$ for each $x^i$, contrary to the work of \cite{noise2noise} which contained 2 realizations of the corrupted images. With this, we come to the description of our manifold.\\
A manifold is a lower dimensional basis for describing a high dimensional data. Images can have a complicated and highly non-linear manifold \cite{Hinton97modelingthe}, \cite{faaeq2018image}, \cite{zhu2016generative}. we attempt to approximately learn this manifold using the noisy instances of the images, since noisy instances will waver around this manifold, with the more noisy instances obviously straying farther from this manifold. we use this intuition to learn a function $f$ that tries to push the noisy instances in the direction of the manifold, so as to recover the clean images $x^i$. Since the noise is random, we can assume it to be around the manifold, shown in Figure \ref{fig:manifold}. One way to have a manifold learning is to have training pairs consisting of clean images with generated synthetic degraded images from the set of clean images, and use a supervised learning approach to learn a mapping from noisy to clean examples. Although that approach works very well in practice, in certain applications, especially medical imaging, one may not have access to a huge dataset of clean training images. In such a scenario, we try to learn the direction to move within the lower dimensional space without actually knowing the line but rather the \textit{normal direction} towards the line. For each noisy example $y^i$ lying somewhere in the lower-dimensional space, we apply the degradation model to the image yielding a more noisy instance $z^i = D(y^i; \sigma^i_r)$ where the choice of $\sigma^i_r$ will be made clear in later sections. The image $y^i$ is closer to the manifold than $z^i$ because the function $D$ cannot recover the original pixel information in the image. For example, for a Gaussian degradation model 
\begin{equation}
    D(x; \sigma) = x + \epsilon  ,  \epsilon \sim  \mathcal{N}(0, \sigma)
\end{equation}
$y^i = x^i + \epsilon^i_1$ and $z^i = y^i + \epsilon^i_2$, which yields $Var(y^i) = (\sigma^i_1)^2$ and $Var(z^i) = (\sigma^i_1)^2 + (\sigma^i_2)^2$ . Thus, $z^i$ has a higher variance and is more corrupted than $y^i$. Similarly for a multiplicative Bernoulli noise, the degradation model is
\begin{equation}
    D(x; q) = m \odot x ,  m_p \sim \text{Bernoulli}(q) 
\end{equation}
where $m$ is a random binary mask, $p$ is the pixel index and $q$ is the probability of the pixel $m_p$ being 1. In this case, $y^i = D(x^i; q^i_{1}) = m^i_{1} \odot x^i$ and $z^i = D(y^i; q^i_{2}) = m^i_{2} \odot y^i$. Hence, $z^i$ can only contain as much pixel information as the pixels from $y^i$, but cannot recover any of the pixels already lost, and hence is farther from the manifold. With this intuition, we attempt to learn a mapping from $z^i$ to $y^i$, where $z^i$ are formed from a random degradation of the already corrupted image $y^i$. we attempt to learn this mapping using a Fully Convolutional Network \cite{long2015fully}, which was used for semantic segmentation and later became popular in literature for image to image mapping tasks. Learning this mapping encourages the network to invert the degradation model by making use of the context around it. To enable using the context, we choose FCN architectures with sufficiently large receptive fields so that the network can utilize as much information as possible. This leads to a training scheme where no clean examples are ever seen by the network. During inference, we feed-forward the images $y^i$ into the network and obtain approximations of the clean image $x^i$. There are 2 methods for performing this inference, as mentioned in later sections. 

\subsection{Loss functions}
One of the factors that comes into play is a choice of the loss function. Both \cite{noise2noise} and \cite{ulyanov2018deep} already show the effectiveness of a loss function for a particular noise removal task. 

\subsubsection{Additive Gaussian noise}
For additive Gaussian noise, the most commonly chosen loss term is the Mean Squared Error (MSE) loss 
\begin{equation}
    Loss(y^i, z^i) = \sum_{i=1}^{M}\sum_{p=1}^{P}(y^i_p - f(z^i; W)_p)^2
\end{equation}
where $W$ are the learnable parameters of the neural network $f$. we also start with MSE loss as a starting loss function for the Additive Gaussian noise. A perceptual loss term (details ahead) is also added to maintain more patch-level details and avoid blurriness in the outputs.

\begin{figure*}
\begin{center}
\includegraphics[width=0.15\linewidth]{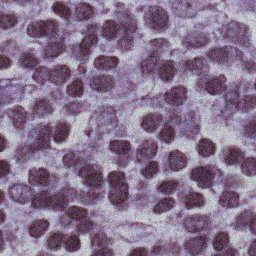}
\includegraphics[width=0.15\linewidth]{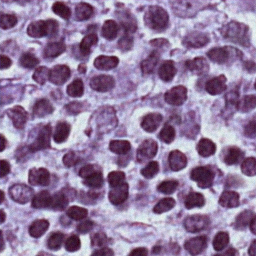}
\includegraphics[width=0.15\linewidth]{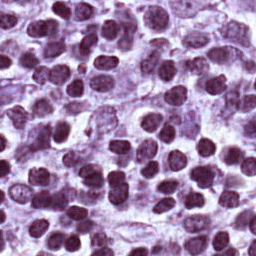}
\hspace{3pt}
\includegraphics[width=0.15\linewidth]{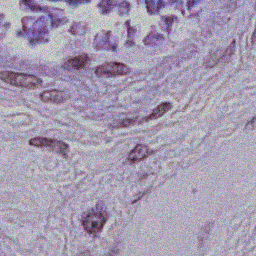}
\includegraphics[width=0.15\linewidth]{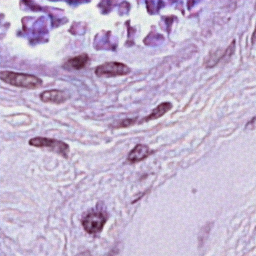}
\includegraphics[width=0.15\linewidth]{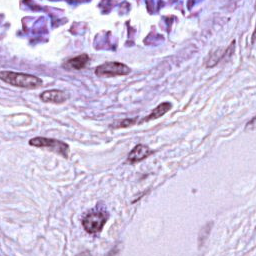}

\includegraphics[width=0.15\linewidth]{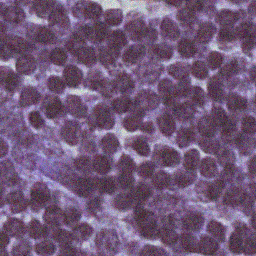}
\includegraphics[width=0.15\linewidth]{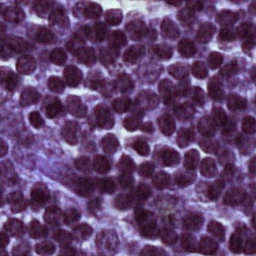}
\includegraphics[width=0.15\linewidth]{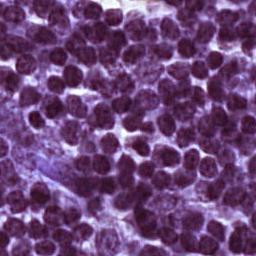}
\hspace{3pt}
\includegraphics[width=0.15\linewidth]{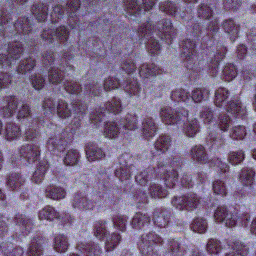}
\includegraphics[width=0.15\linewidth]{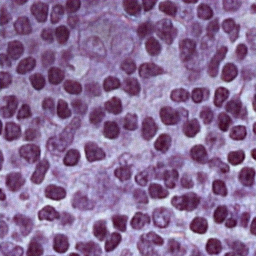}
\includegraphics[width=0.15\linewidth]{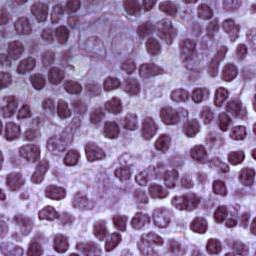}

\includegraphics[width=0.15\linewidth]{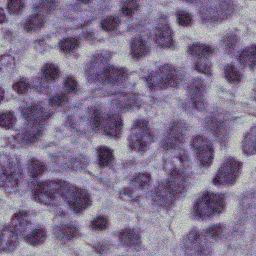}
\includegraphics[width=0.15\linewidth]{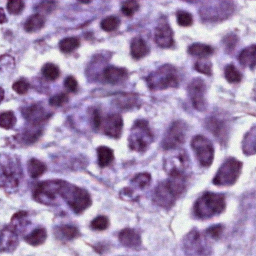}
\includegraphics[width=0.15\linewidth]{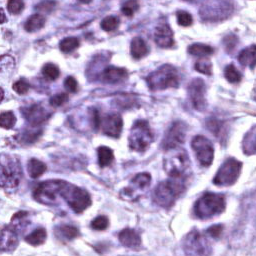}
\hspace{3pt}
\includegraphics[width=0.15\linewidth]{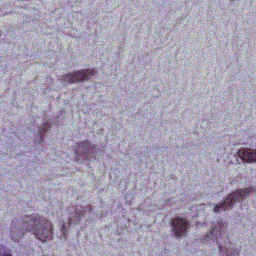}
\includegraphics[width=0.15\linewidth]{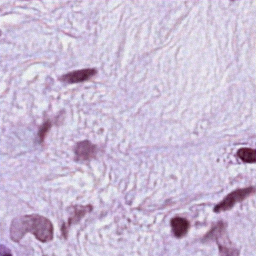}
\includegraphics[width=0.15\linewidth]{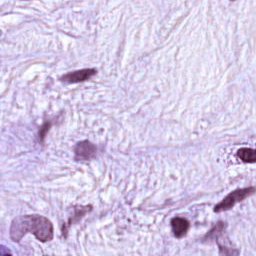}

\end{center}
  \caption{Results on the Camelyon 16 dataset. The images from left to right are (i) Noisy image (ii) Reconstructed image, and (iii) original image. The network removes the Gaussian noise and retains most of the crispness of edges across a wide variety of validation images.}
\label{fig:camelyon}
\end{figure*}

\subsubsection{Multiplicative Bernoulli Noise}
For multiplicative Bernoulli noise, the loss function is modified so that the pixels which are missing do not count towards the loss function. The loss term is modified from the MSE loss function as:
\begin{equation}
    Loss(y^i, z^i) = \sum_{i=1}^{M}\sum_{p=1}^{P}(m_p\odot(y^i_p - f(z^i; W)_p))^2
\end{equation}
where $m$ is a binary mask indicating whether the pixel is missing or not. This can be obtained simply by checking which pixels are zero-valued.

\subsubsection{Poisson Noise}
For Poisson noise, we use the MSE loss function as well, since the value of the pixel is same in expectation. Although $z^i$ does not maintain a Poisson distribution when $z^i$ is formed from $y^i$, we hypothesize that the output of the neural network $f(z^i; W)$ and $y^i$ would belong to the same distribution (since the network is encouraged to learn it). Hence, the network is expected to invert the degradation model. If there existed a deterministic function $g$ such that $y = g(x)$ and $z = g(y)$ and the network $f$ was a good approximation of $g^{-1}$, then for a particular $y$ the network would output $f(y) \approx g^{-1}(y) = x$. The degradation model isn't deterministic, but over a long number of training iterations, the model should figure out to output an average output of a particular input. Therefore, a MSE loss term seems like a good baseline loss function to train on. MSE loss will ensure that $y^i$ will be of the same distribution as $f(z^i; W)$ as the model learns to invert the Poisson degradation. To preserve higher level information, we also add a Perceptual Loss term.

\subsection{Perceptual Loss}
To compensate for the problems associated with bluriness due to the averaging effect of MSE loss and loss of perceptual features, \cite{johnson2016perceptual}, \cite{ledig2017photo} propose an additional perceptual loss term which also maps higher order relationships between the 2 images. In my case, the perceptual loss only contains of a content loss given by:
\begin{equation}
    Loss(y^i, z^i) = \sum_{i=1}^{M}\sum_{p'=1}^{P'}(\phi(y^i)_p' - \phi(f(z^i; W))_p')^2
\end{equation}
where $\phi : \mathbb{R}^{H\times W} \longrightarrow \mathbb{R}^{H'\times W'}$ is a mapping from the image to a feature map, usually the output of an intermediate layer of a network pretrained on thousands of images. In my case, $\phi$ is the output of the \textit{relu2\_2} of a VGGNet \cite{simonyan2014very} trained on the Imagenet \cite{imagenet_cvpr09} dataset. Since I'm training on medical datasets and we use a network pretrained on a very different dataset, we use the outputs of an early layer, which usually only gives a combination of low-level information like gradients, texture, etc. In my experiments, we see that adding a Perceptual loss term helps in removing the blurriness of the outputs. 

\subsection{Iterative refinement}
Given the training scheme, the inference can be done simply by passing the noisy image through the network and getting the output. However, during training, the network sees a lot of variation between the input and outputs, as multiple instances of $z^i$ are created for a particular instance of $y^i$ for training. A particular value of $z^i$ may be close to one or more instances of $y^i$ in the manifold. Hence, the network may take sub-optimal steps in order to minimize the net error across all training examples. To tackle such a case, we take the output of the network, take a weighted average of the current output and previous output and pass it again as input for the next iteration. Although this makes the inference time more than the current baseline, we observe subtle improvements in the validation scores. The algorithm is outlined as follows:
\begin{algorithm}
\SetAlgoLined
\KwResult{Network output $out$}
 $out = f(y; W)$\;
 $iter = 0$\;
 \While{$iter <  NumIters$}{
  $tmpout = f(out; W)$\;
  $out = (1-\alpha)\cdot out + \alpha\cdot tmpout $\;
  $iter = iter + 1$\;
 }
 \caption{Iterative refinement}
\end{algorithm}
The parameters $\alpha$ and $NumIters$ can be chosen to fit a validation dataset. we observed that the optimal values of these parameters depend on the noise levels as well as type of noise. The update term can also be seen as $out + \alpha\cdot(tmpout - out)$ which corresponds to a residual update with learning rate $\alpha$. we describe some implementation details in the next section.

\begin{figure*}
\begin{center}
\includegraphics[width=0.15\linewidth]{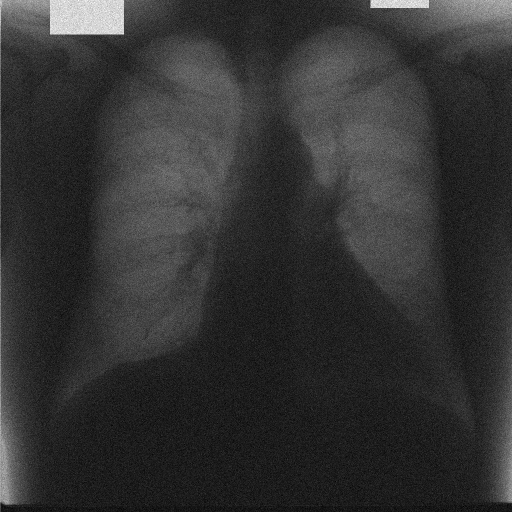}
\includegraphics[width=0.15\linewidth]{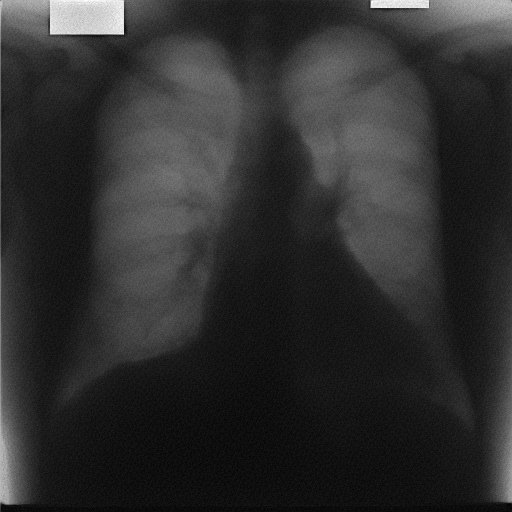}
\includegraphics[width=0.15\linewidth]{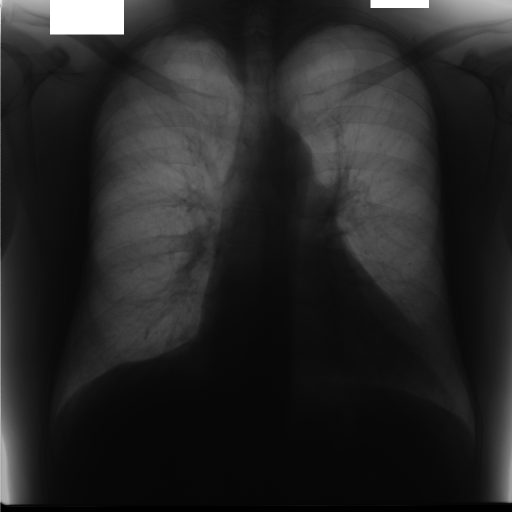}
\hspace{3pt}
\includegraphics[width=0.15\linewidth]{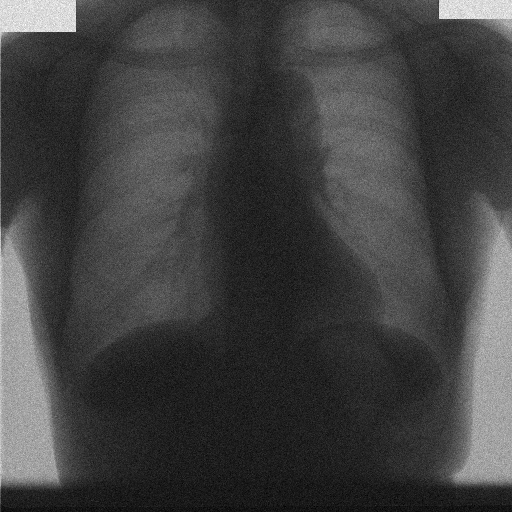}
\includegraphics[width=0.15\linewidth]{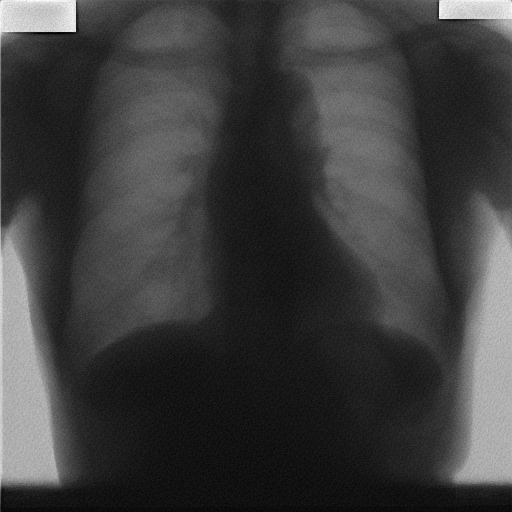}
\includegraphics[width=0.15\linewidth]{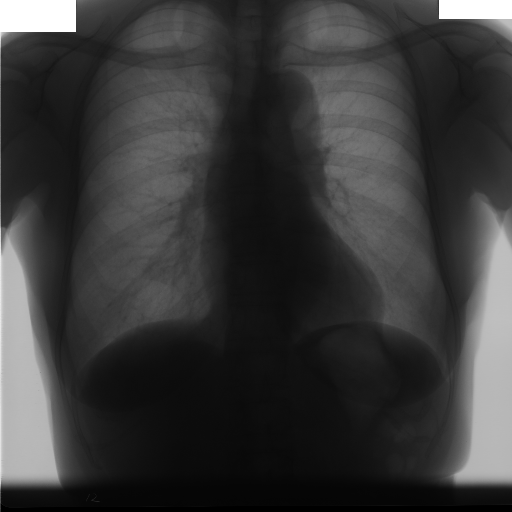}

\includegraphics[width=0.15\linewidth]{images/poisson/ni1.png}
\includegraphics[width=0.15\linewidth]{images/poisson/ci1.png}
\includegraphics[width=0.15\linewidth]{images/poisson/im1.png}
\hspace{3pt}
\includegraphics[width=0.15\linewidth]{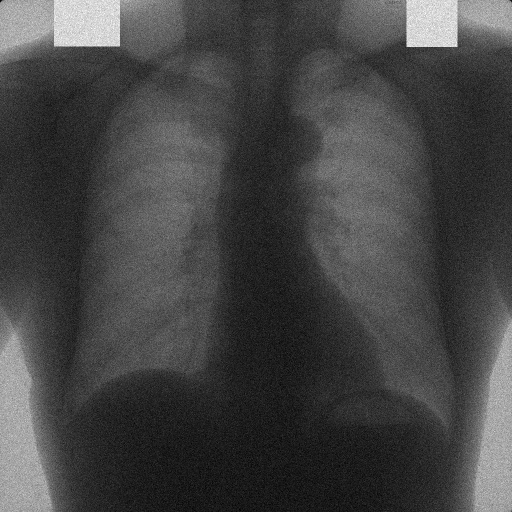}
\includegraphics[width=0.15\linewidth]{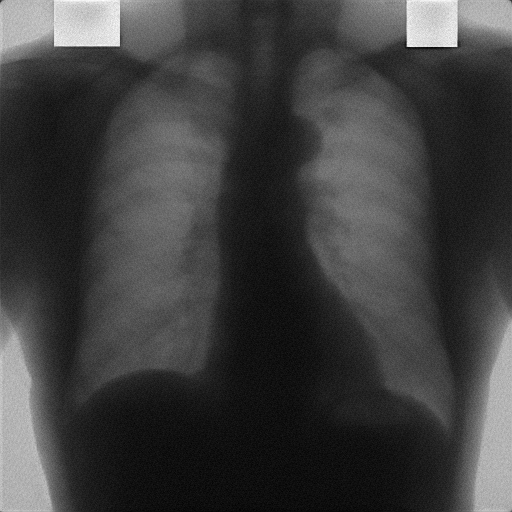}
\includegraphics[width=0.15\linewidth]{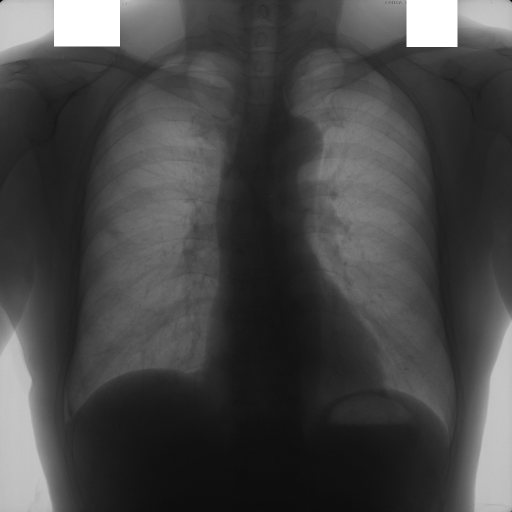}
\end{center}
  \caption{Results on the JSRT validation dataset. The images from left to right are (i) Noisy image (ii) Reconstructed image, and (iii) original image. The network removes most of the graininess caused by the Poisson noise, while preserving the important parts of the image. Note that there is still a little graininess in the image, we postulate that this is because the Poisson model isn't additive or multiplicative, and hence is a little harder to remove without seeing any clean examples.}
\label{fig:jsrt}
\end{figure*}

\section{Implementation}
For all the experiments, we use the commonly used U-Net architecture \cite{ronneberger2015u}. All networks are trained with the Adam optimizer \cite{kingma2014adam}, with a learning rate of $10^{-4}$ and weight decay of $10^{-5}$. All the inputs are scaled to the range of $[0, 1]$. we use 3 medical datasets for different tasks to show the effectiveness of the method over a variety of datasets. In particular, we use 2D slices of T1-weighted images from a subset of the IXI Guys dataset \cite{IXI} containing 322 3-dimensional T1-weighted images. we divide the dataset into 289 images for training, and 33 images for validation. For training, we modify the dataset as following:  All the noisy images $y^i$ were generated by retaining every pixel of $x^i$ with a probability $q^i_1$. The value of $q^i_1$ is chosen from the range of $[0.35, 0.85]$. For each $y^i$, an instance $z^i$ is created by further training the pixels of $y^i$ with a probability $q^i_2$. The value of $q^i_2$ is chosen randomly between $[0, 0.8]$. During validation, all the images are corrupted with Bernoulli noise of probability $q = 0.5$, and metrics were compared with the ground truth slices. Perceptual loss was \textit{NOT} used in this case, since Perceptual loss would also capture the patch-level similarity, and incorporating the binary mask in this loss would be non-trivial. Finally, the iterative refinement is applied for 10 iterations with $\alpha = 0.01$.  

For the experiment with Gaussian noise, we use the Camelyon 16 dataset \cite{Camelyon}, which contains whole-slide images of hematoxylin and eosin (H\&E) stained lymph node sections. Specifically, we use $256\times256$ patches from whole-slide images for the training and validation sets. The training set consists of 6100 patches uniformly chosen from the training set, and 1200 patches chosen uniformly from the validation set. The clean images $x^i$ are corrupted with Gaussian noise with iid Gaussian noise of standard deviation $\sigma^i_1$. The value of $\sigma^i_1$ is randomly chosen from 3 to 50 pixels on a scale of 255 pixels for the image. The image intensities are scaled to the range $[0, 1]$, and this forms a particular instance of $y^i$. From this instance, a more noisy version $z^i$ is created by adding iid Gaussian noise to $y^i$ with standard deviation $\sigma^i_2$ ranging from 0 to 25 pixels. During validation, the standard deviation of $y^i$ is always fixed to 20 pixels. The perceptual loss term is used along with the MSE term. A scaling factor of 0.5 suffices in this case. The iterative refinement is applied for 10 iterations with $\alpha$ set to 0.01.

For the experiment with Poisson noise, we chose the JSRT dataset \cite{jsrt}. The dataset contains $2048\times2048$ sized high-resolution chest X-ray images. we follow the 50-50 train-val split as prescribed by \cite{jsrt}. However, in the interest of faster computation, the images are resized to a size of $512\times512$ pixels. For a given $x^i$, $y^i$ is generated by applying Poisson noise to the clean image. Furthermore, $z^i$ is generated by applying Poisson noise to $y^i$. Note that $z^i$ doesn't follow the same distribution as $y^i$ (which was the case in the previous cases). However, $z^i$ was found to be more noisy than $y^i$ in my experiments, and we decided to choose MSE + Perceptual loss as the loss function as a starting baseline. The coefficient of perceptual loss was set to 0.2 for my experiments with this dataset. The training is done for 30 epochs. Iterative refinement was applied during testing for 100 iterations with $\alpha$ set to 0.01.

Note that in all these cases, only a single $y^i$ is formed from a given $x^i$, so that the model doesn't see multiple instances corresponding to a single $x^i$ and learn to predict it so as to minimize the average loss, as opposed to the code released by \cite{noise2noise}, where different noisy instances are computed every iteration for a given clean image. Since $z^i$ are always different, overfitting was not a concern in my experiments. 

\section{Results and Discussion}
We use the 2 commonly used metrics for denoising: (i) PSNR (Peak Signal-to-noise ratio) and (ii) SSIM (Structural Similarity index). PSNR is generally used as a quality measurement between a good image and a compressed image \cite{psnr}. PSNR is also a commonly used metric for measuring quality between denoised image and a good quality image as well. PSNR for 2 images $x$ and $y$ is defined as 
\begin{equation}
    PSNR(x, y) = 10\log_{10}{\Big(\frac{R^2}{MSE(x, y)}\Big)}
\end{equation}
where $MSE(x, y)$ is the mean squared error between the predicted clean image and the ground truth image. $R$ is the maximum fluctuation in the input image data type. In this case, since we scale the image from 0 to 1, $R = 1$. The second metric, SSIM is a metric used to measure structural similarity between 2 images based on local means and variances. SSIM for 2 images $x$ and $y$ is defined as  
\begin{equation}
    SSIM(x, y) = \frac{(2\mu_x\mu_y + c_1)(2\sigma_{xy} + c_2)}{(\mu_x^2 + \mu_y^2 + c_1)(\sigma_x^2 + \sigma_y^2 + c_2)}
\end{equation}
where $\mu_x, \mu_y$ are the averages of $x, y$, $\sigma_x, \sigma_y$ are the standard deviations, $\sigma_{xy}$ is the covariance of $x$ and $y$. $c_1$ and $c_2$ are constants chosen to be $(0.01)^2$ and $(0.03)^2$ respectively (which are the default values). 

\subsection{Bernoulli noise}
Figure \ref{fig:ixiguys} shows some input images, their reconstructions, and the ground truth image. Unlike the experiments done in \cite{noise2noise}, my training pairs do not contain any extra information other than $y^i$. This makes the problem harder for the neural network, since it has to fill up the points itself without mutual sharing of information between noisy pairs. However, we see that the network learns to effectively reconstruct the entire image with great accuracy.

\begin{figure}
    \centering
    \includegraphics[width=0.3\linewidth]{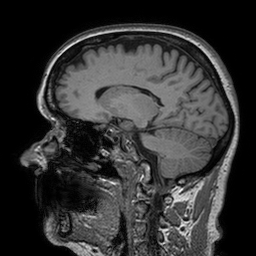}
    \includegraphics[width=0.3\linewidth]{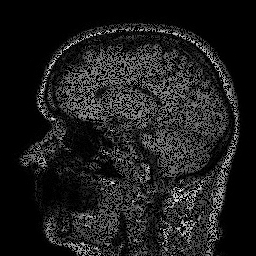}
    \includegraphics[width=0.3\linewidth]{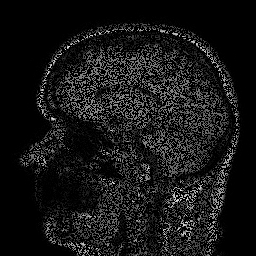}
    \caption{Original patch ($x^i$), noisy version ($y^i$), and more corrupted version $z^i$ from the IXI Guys dataset. Note that only $y^i$ and $z^i$ are used during the training process}
    \label{fig:ixidemo}
\end{figure}

\begin{table}
\begin{center}
\begin{tabular}{|l|c|c|}
\hline
  & SSIM & PSNR \\
\hline\hline
Noisy & 0.5029 & 14.2522 \\
Ours & \textbf{0.7363} & \textbf{32.3628}\\
\hline
\end{tabular}
\end{center}
\caption{Results on the IXI Guys validation set.}
\label{table:ixi}
\end{table}

Training was done for 15 epochs on a NVIDIA GTX 1080 GPU. Notable performance was recorded in terms of improvement in SSIM and PSNR. 

\subsection{Gaussian Noise}
Figure \ref{fig:camelyon}
shows some input images, their reconstructions, and the ground truth image, similar to the previous experiment. Here, the network is encouraged to learn less noisy versions of the images, and the network performs well on the validation dataset, owing to the variation in the noise that the model gets to see throughout the training process.

\begin{figure}
    \centering
    \includegraphics[width=0.3\linewidth]{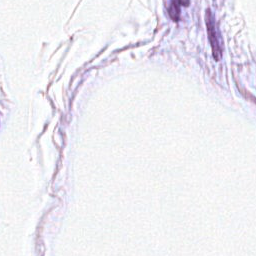}
    \includegraphics[width=0.3\linewidth]{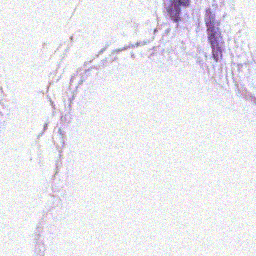}
    \includegraphics[width=0.3\linewidth]{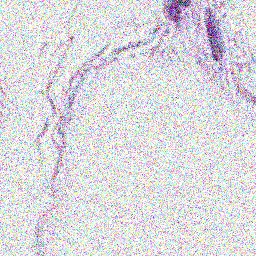}
    \caption{Original patch ($x^i$), noisy version ($y^i$), and more corrupted version $z^i$ from the Camelyon 17 dataset. Note that only $y^i$ and $z^i$ are used during the training process}
    \label{fig:camelyondemo}
\end{figure}
Here, the model can make use of the Perceptual loss term as well, to exploit more patch-level similarities and correspondences. Training the network with and without the perceptual loss term shows a difference in the validation scores when trained for the same number of epochs. In particular, the perceptual loss reduces blurriness of the output and maintains crisper edges than the only-MSE loss based variant. Table \ref{table:camelyon} highlights the difference. Some qualitative results can also be seen in Figure \ref{fig:camelyon}.

\begin{table}
\begin{center}
\begin{tabular}{|l|c|c|}
\hline
  & SSIM & PSNR \\
\hline\hline
Noisy & 0.7102 & 26.1487 \\
BM3D & 0.8518 & 29.2043 \\
Ours (MSE) & 0.8522 & 29.2364 \\
Ours (MSE + Perceptual) & \textbf{0.8931} & \textbf{30.1610}\\
\hline
\end{tabular}
\end{center}
\caption{Results on the Camelyon validation set.}
\label{table:camelyon}
\end{table}

The network can remove most of the noise without distorting or blurring the features in the image. Random images are picked from the validation set to demonstrate the effectiveness of this training method. 

\subsection{Poisson Noise}

\begin{figure}
    \centering
    \includegraphics[width=0.3\linewidth]{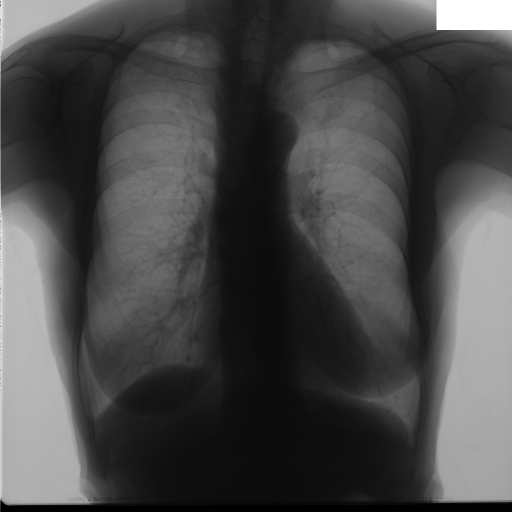}
    \includegraphics[width=0.3\linewidth]{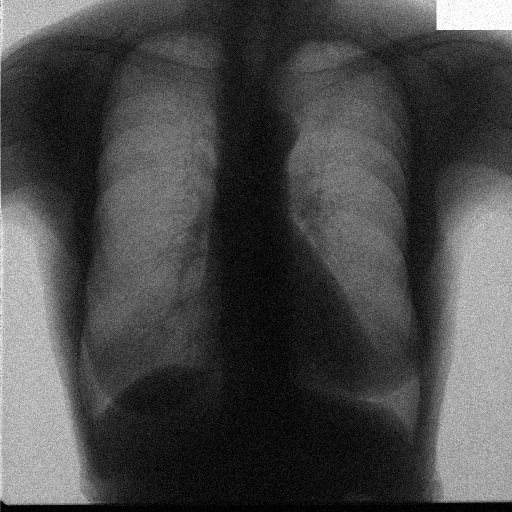}
    \includegraphics[width=0.3\linewidth]{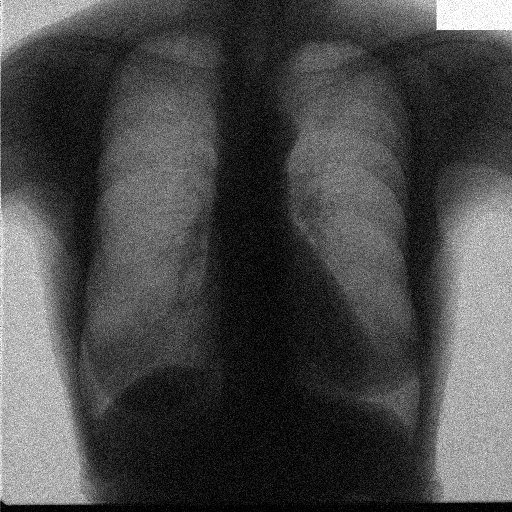}
    \caption{Original patch ($x^i$), noisy version ($y^i$), and more corrupted version $z^i$ from the JSRT dataset. Note that only $y^i$ and $z^i$ are used during the training process (like the previous experiments)}
    \label{fig:jsrtdemo}
\end{figure}

Figure \ref{fig:jsrt} shows some of the results of denoising on X-ray images corrupted by Poisson noise. This dataset was particularly hard because the images do not have a very high contrast. The ribs and other structures are barely visible. Also, the images $y^i$ and $z^i$ do not follow the same distribution unlike the other cases, so the task is expected to be harder than the previous tasks. However, MSE loss seems like a good baseline and can reduce sufficient amount of noise from the corrupted image. Table \ref{table:jsrt} shows the difference between noisy and denoised images. 

\begin{table}
\begin{center}
\begin{tabular}{|l|c|c|}
\hline
  & SSIM & PSNR \\
\hline\hline
Noisy & 0.5253 & 24.2515 \\
Ours & \textbf{0.6692} & \textbf{27.4641}\\
\hline
\end{tabular}
\end{center}
\caption{Results on the JSRT validation set.}
\label{table:jsrt}
\end{table}

There is an improvement from the baseline noisy images. Note that training is done by corrupting the noisy instances from the training set to generate training pairs $y^i, z^i$, and no information about clean images is known. Both PSNR and SSIM show significant improvements. Although the approach isn't as good as training with clean samples, there are non-trivial improvements and it is still comparable to fully supervised alternatives. \\

\section{Conclusion}
We proposed a method of using a dataset containing \textit{only} noisy examples, and a training strategy to make the model infer the dynamics of the particular noise model. The denoising is \textit{blind} in the sense that the parameter of the noise model is not known or can be different for different images. We show significant denoising achieved in 3 different datasets, and comparable performance with other supervised algorithms. Owing to the power of deep neural networks, we show that neural networks can be made into good approximators of stochastic degradation models by trying to push images into the manifold of clean images. This interpretation of trying to learn a manifold and hence perform  learning is a neat technique when there is an unavailability of clean data, which is more common in medical settings. \\

{\small
\bibliographystyle{ieee}
\bibliography{egbib}
}

\end{document}